# Graph-of-Causal Evolution: Challenging Chain-of-Model for Reasoning

Libo Wang

UCSI University

free.equality.anyone@gmail.com

## Abstract

In view of the problem that each subchain in the chain-of-model (CoM) relies only on the information of the previous subchain and may lose long-range dependencies due to the causal mask blocking the global context flow between multi-level subchains, this work proposes a graph of causal evolution (GoCE). Its core principle is to map the implicit token representation into a differentiable and sparse causal adjacency matrix, then permeate causal constraints through each layer of calculation using causal-masked attention and causal-MoE. By combining intervention consistency loss test and self-evolution gate, the dynamic balance between causal structure learning and adaptive updating of transformer architecture is realized. The researcher built experimental environments in sandboxes built with Claude Sonnet 4, o4-mini-high, and DeepSeek R1 respectively with the transformer variant architecture introduced in GoCE. It is evaluated on publicly available datasets including CLUTRR, CLADDER, EX-FEVER, and CausalQA and compared with the baseline LLMs. The finding proves that GoCE strengthens the transformer's ability to capture long-range causal dependencies, while the ability to self-evolve is improved. It not only surpasses the design of CoM in terms of design principles, but also provides experience for future research on causal learning and continuous adaptive improvement.

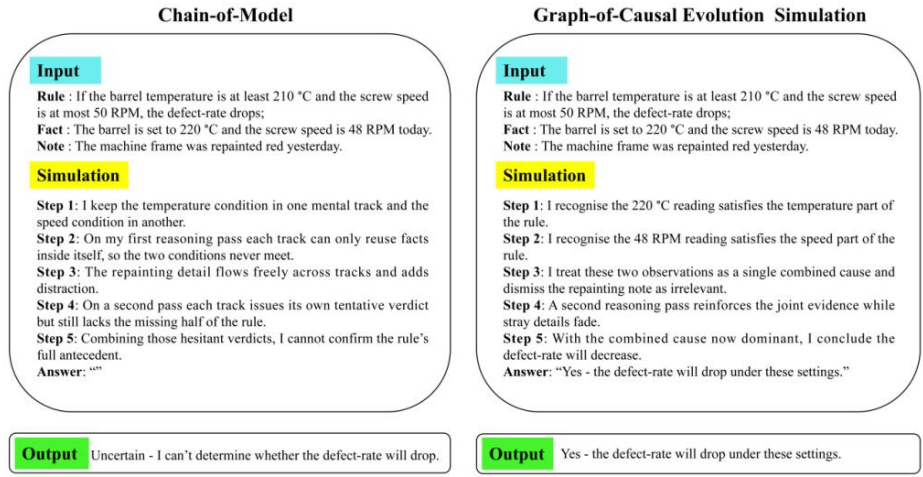

Figure 1: Comparison of the reasoning of chain-of-model and graph-of-causal evolution via the example.

## 1. Introduction

As the technology in the field of reasoning matures, large language models (LLMs) have evolved from simply predicting the next word to a complex system that deeply combines attention heads (Tikhonov & Ryabinin, 2021; Zhang et al., 2023; Chen et al., 2024). The self-attention mechanism calculates the correlation strength between different representation dimensions and the multi-head structure captures the complex representation of diverse semantic subspaces, laying the foundation for maintaining multi-granular semantic consistency during implicit reasoning (Vaswani et al., 2017; Zheng et al., 2024). The introduction of the Chain-of-thought (CoT) technology aims to create multi-step intermediate semantic nodes in the input prompt to explicitly guide the model's hidden layer to reversely infer the premise from the conclusion to overcome the insufficient capture of long-distance dependencies in pure direct output (Wei et al., 2022;

Diao et al., 2023; Mitra et al., 2024). Notably, CoT simulates reasoning steps in the prompt phase, it is difficult to establish a persistent causal structure within the model (Yu et al., 2023; Yu et al., 2024).

Given the lack of multi-level causal dynamic scheduling in the hidden representation of the transformer layer in reasoning tasks, Song et al. (2025) proposed chain-of-representation (CoR) that divides each layer of hidden representation into multiple sub-representations (chains) with causal relationships. Based on the characteristics of this concept, chain-of-layer (CoL) reconstructs the network layer so that the output of each sub-chain depends only on the previous input. It means that the model introduces causal dependencies between different scales to ensure that each scale can only use the information of the previous scale, and finally forms a chain-of-model (CoM) learning paradigm at the model level.

However, the problem is that causal mask blocks the global context flow between multi-level subchains, which means that each subchain only relies on the information of the previous subchain and it is difficult to integrate the complete semantics in parallel, so long-term dependencies may be lost. It relies on random or truncated initial weights and does not provide a smooth mapping function when adding subchains, which exacerbates the fragmentation of the representation space and increases the difficulty of retaining existing knowledge during expansion. In the face of multi-scenario reasoning requirements, the causal mask of the subchain only transmits information along a single path each time, which may eliminate the parallel computing advantages of multi-layer latent representations in practice. Although CoLM-Air attempts to reduce pre-filling overhead by sharing KV cache, it does not consider the significant differences in the distribution of latent vectors of different subchains (Song et al., 2025).

Researchers claim that CoM claims multi-level subchain operations while blocking cross-chain information with causal masks, just like building a thousand-story building but removing all the ladders. The gap is that CoM forces each layer of latent representation to be split and only allows the i-th level subchain to access the 1st to $i$-1 level outputs, and it only retains the key-value mapping of the preceding subchain in the attention calculation, which eliminates the possibility of cross-group parallel fusion. This design prevents subsequent subchains from sharing the same-level and higher-level intelligence instantly, resulting in the long-range semantics required for each layer having to iterate along the coherent path. It is difficult for LLMs to absorb global features in parallel and maintain semantic coherence during multi-hop reasoning or ultra-long context processing. Besides, multi-head attention and block position encoding are difficult to work together because the mask restriction only transmit information along a single path.

To design a feasible solution for the gap, this work proposes "Graph-of-Causal Evolution (GoCE)" to replace the linear subchain of CoM. To address the core gap of global context disconnection caused by causal masks, it uses differentiable sparse gating to reconstruct cross-layer connections so that each node can capture the complete semantics in parallel instead of sequentially. It uses a dynamic edge weight update mechanism to fine-tune the weights between nodes based on gradients and historical distributions to ensure that new nodes can smoothly inherit existing representations in the shared embedding space. In addition, GoCE freezes stable nodes through self-evolving routing and dynamically adjusts paths according to scenario requirements, achieving continuous maintenance of long-range dependencies and multi-scenario flexible reasoning.

## 2. Related Work

The previously mentioned Song et al. (2025) proposed "Chain-of-Model (CoM)" as a learning paradigm that splits the hidden vector into multiple "Chain-of-Representation (CoR)" sub-vectors. It ensures that the output of the i-th chain depends only on the previous $i$ chains of the input hidden representation through the "Chain-of-Layer (CoL)" mechanism, thereby introducing explicit causality and composability.

The CoM emphasizes the introduction of multi-chain splitting based on the dimension ratio $C=\{c_1,…,c_n\}$ in the linear, attention, feed-forward network and regularization modules of each layer of transformer, so that the output of the i-th chain only calculates the sub-matrix $W_i$ and bias corresponding to the input of the previous $i$ chains ($x \leq i$). The CoM forces the query, key, value and output to be rewritten in chain-of-linear, and maps the number of attention heads to each chain according to $C$ to ensure that each chain only processes single-scale information. In addition, it also proposes KV sharing technology that calculates all keys and values only on the first chain and copies them to subsequent chains to achieve seamless switching and pre-filling acceleration between different sub-models (Song et al., 2025).

This work adopts the concept of "graph" instead of the traditional "chain" due to the need for multi-dimensional structural information and nonlinear reasoning capabilities of LLMs. As shown in the study of Dwivedi and Bresson (2020), the graph transformer uses a neighborhood-aware attention mechanism to make each node calculate attention weights only with its neighboring nodes. Yao et al. (2023) also pointed out in their research on chain-of-thought that the human thinking process is not a simple linear continuous chain of thought, but rather a jumpy, non-continuous thought graph. Researchers found that it is difficult for LLMs to simulate the real thinking process by relying solely on chain-of-thought, so they designed graph-of-chain (Yao et al., 2023).

In addition, Zhang et al. (2025) proposed to realize the self-improvement ability of artificial intelligence based on Gödel machine and automatically discover new algorithms through meta-learning. This work draws on the concept of self-improvement. Shafayat et al. (2025) use online self-training reinforcement learning algorithm to use the self-consistency of the model to infer the correctness signal and train without any real value supervision (Shafayat et al., 2025).

## 3. Graph-of-Causal Evolution

As a self-evolving causal reasoning paradigm, Graph-of Causal Evolution (GoCE) aims to ensure causal dependencies in the reasoning process and continuously update the structure in multiple rounds of reasoning. Compared with CoM, it dynamically builds and iteratively evolves the causal graph through text prompts, so that LLMs can continuously correct their own causal edge weights to achieve more accurate intervention and counterfactual reasoning. Figure 1 shows the reasoning steps of simulating the answer to the example based on the principles of CoM and GoCE respectively by showing the solution.

### 3.1 Latent-Causal Graph Builder

The components of are closely connected to each other, and together they complete the construction of causal implicit representations from text input based on the transformer architecture (Figure 2).

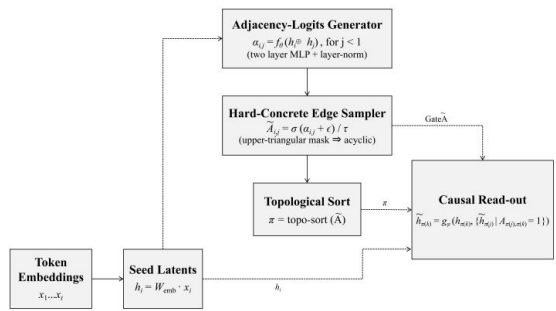

Figure 2: Latent-Causal Graph Builder

The token embeddings are responsible for mapping each word into a continuous vector of fixed dimension, which serves as the basic representation for subsequent processing. Next, the seed latents treat the embedding vectors as latent nodes. This component provides a set of initial latent states to carry the reasoning of the subsequent causal structure. Entering the adjacency-logits generator, the system compares each pair of latent nodes through TinyMLP to generate a score value representing the strength of the potential causal connection, which can be interpreted as the probability of the existence of a hypothetical edge to pave the way for the next step of discretization. The hard-concrete edge sampler converts continuous scores into discrete edges using a differentiable gating technique, which ensures that the constructed causal graph is sparse and acyclic while maintaining the advantages of differentiability.

After discretization, the topological sort generates a node processing order by topologically sorting the generated acyclic neighbor structure to avoid causal loops and provide a clear path for subsequent causal information reading. The causal read-out reads each node one by one according to the topological order. It combines the original latent vector of the current node with the updated latent representations of all its parent nodes, and fuses the information from all parties through a specially designed feedforward network to generate a new causal latent representation. Finally, the output aggregates the updated latent vector and outputs it to the subsequent transformer layer or the final classifier to achieve seamless transmission of implicit causal information in the architecture. The corresponding algorithm is as follows:

$$\tilde{h} = g_\phi \left( h_i, \sum_{j<i}[\text{HardConcrete}_\tau(f_\theta(h_i, h_j))] h_j \right)$$

where $h_i/h_j$ represent seed latent vector; $f_\theta$ represents that two-layer MLP produces one scalar edge logit; $\tilde{h}_i$ represents causally refine latent forwarded to downstream modules; $j<i$ represents the guarantees acyclicity.

## 3.2 Causal-Masked Multi-Head Attention

The core goal of this module is to use the previously generated topologically ordered latent representations and dynamic causal structures to complete weighted aggregation (Figure 3).

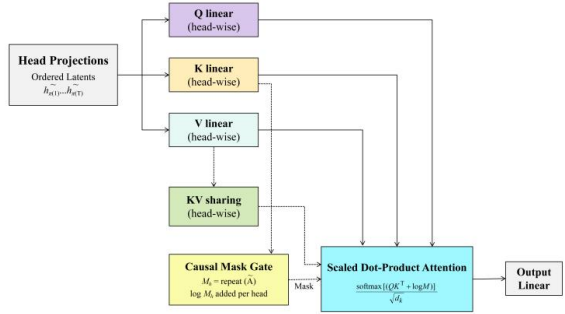

Figure 3: Causal-Masked Multi-Head Attention

First, the head projections receive the ordered latents output from the causal read-out stage, which represent the causal topological order and ensure that downstream calculations can follow this order and structure prior. The head projections prepare independent query ($Q$), key ($K$) and value ($V$) basis for multi-head attention through the head projection mechanism, which enables different attention heads to focus on different subspaces of latent vectors. $Q$ linear, $K$ linear and $V$ linear perform linear mapping on the ordered latent representations, respectively, and produce the corresponding $Q$ matrix, $K$ matrix and $V$ matrix.

The $K$ linear and $V$ linear are introduced into the $KV$ Sharing mechanism that is responsible for copying the key and value vectors to all attention heads, which realizes one-time calculation and multiple uses. After obtaining the value vectors of $Q$, $K$ and $V$, the causal mask gate repeatedly expands the previously generated binary causal graph neighbor matrix $\tilde{A}$ to match the size of each attention head, forming a head-wise mask matrix represented as $M_h$. The mask is integrated into the attention score in the form of log $-\infty$ or 0, so that the model only allocates attention to parent nodes that meet the causal order and strictly isolates interference from non-causal paths. When entering the scaled dot-product attention stage, the system calculates the initial similarity between the $Q$ vector and the $K$ vector through the inner product, and then combines it with the mask provided by the causal mask gate, and weightedly aggregates it with the $V$ vector after softmax to generate the intermediate output of each head. The output linear concatenates or sums the output vectors of multiple attention heads and passes through the final linear projection.

### 3.2.1 Causal-Sparse Attention with Interventional Loss (CSAIL)

The CSAIL algorithm achieves the coordinated optimization of attention and routing mechanisms by integrating causal masks and intervention consistency loss. Specifically, it maps the causal graph structure to a mask matrix in multi-head attention, so that each $Q$ can only interact with its causal predecessor or successor nodes, thereby enforcing the causal order obtained by the previous topological sorting in the latent dimension. Only then does it share the same $K$ and $V$ representations for each attention head, while avoiding redundant calculations and ensuring that different heads focus on the same causal region. Through the top-k router, only the pre-activated expert nodes are activated in the causal neighborhood to dynamically select the most representative nodes to participate in reasoning. The details are as follows:

$$H \in \Re^{T \times d}, W_Q, W_K, W_V \in \Re^{d \times d_k}, A \in \{0,1\}^{T \times T}, g_\eta, \lambda_{L_0}, \tau_{cf}$$

$$F_{\text{CSAIL}}(H, A; \Theta) = \operatorname{argmin}\{L_{\text{pred}}(Z) + \lambda_{L_0}(\|W_Q\|_0 + \|W_K\|_0) + \text{KL}[\text{softmax}(\frac{QK^T + \ln A}{\sqrt{d_k}}) \| \text{softmax}(\frac{QK^T + \ln A}{\sqrt{d_k}\tau_{cf}})]\}$$

$$Z = \text{soft max}(\frac{QK^T + \ln A}{\sqrt{d_k}})V$$

$$E_t = \text{Top-}k(g_\eta(h_t), k, \text{mask} = E_{v_t})$$

$$v_t = \{j \mid A_{tj} = 1 \text{ or } A_{jt} = 1\}$$

where KL term is the intervention consistency loss; $\tau < 1$; $\Theta$ collects $\{W_Q, W_K, W_V, \eta\}$; $\lambda_{L0}$ is the small hyperparameter weighting the $l_0$ sparsity penalty; $\tau_{cf} \in (0,1]$ represents the counterfactual temperature.

Finally, it calculated the two sets of attention distributions during the training and intervention phases, and measured their differences using KL divergence as an intervention consistency loss.

**3.3 Causal-Conditioned Sparse-Expert Feed-Forward**

In the first path shown in Figure 4, the attended latents are sent to the causal router. It corresponds to each expert through the pre-trained routing vector, and generates a set of expert preference logits for each moment latent vector to measure which experts are suitable for the current latent representation to be processed by (Zhou et al., 2022; Cai et al., 2025). The causal directed acyclic graph (DAG) enters the vicinity mask. It assigns the node at that moment and all its directly connected causal neighbors to the causal neighbor domain set $v_t$ to represent the scope of experts that can participate in reasoning (Lipsky & Greenland, 2022; Zeng et al., 2025).

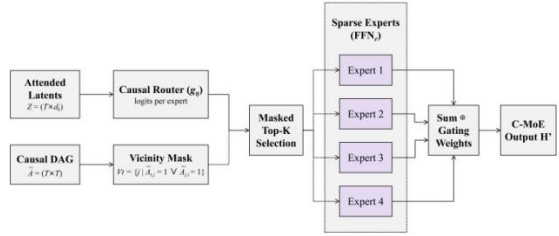

Figure 4: Causal-Conditioned Sparse-Expert Feed-Forward

Next, the masked top-k selection receives both the original scores of the causal router and the set of allowed experts generated by the vicinity mask. It only retains the original values of the scores that meet the causal neighborhood conditions, and the other scores are masked to extremely low values. Next, the system selects the best experts from the unmasked scores as the sparse feedforward subnetwork. In the sparse experts, the selected $expert_1$ to $expert_n$ each perform a small feedforward network calculation for the implicit vector at that moment, while other unselected experts do not participate in the calculation to save resources. The sum $\oplus$ gating weights is responsible for summing the outputs of all selected experts according to the gating weights. Since $k$ is usually set to 1, it is equivalent to directly adopting the output of the expert. Finally, the C-MoE Output $H'$ takes the above summation result as the updated implicit representation of this moment and passes it to the subsequent layer or downstream task. The corresponding algorithm is as follows:

$$\forall t \in [1, T]: \ \text{logit}_{t,e} = g_\eta(h_t, e) \cdot 1\{e \in Ev_t\},$$
$$E_t = \text{Top-k}(\text{logit}_{t,\cdot}, k),$$
$$h'_t = \sum_{e \in E_t} \sigma(\text{logit}_{t,e}) \text{FFN}_e(h_t)$$

where $h_t$ is latent token vector (row of $Z$); $g_\eta(h_t, e)$ represents router network produce a real logit for expert $e$; $E_{vt}$ represents experts that have been already active in $t$'s causal neighbourhood; $E_t$ is final expert set for $t$ after masked top-k; $\text{FNN}_e$ is feed-forward weights of expert $e$.

**3.4 Intervention & Counterfactual Module**

The module is dedicated to verifying and guiding the consistency of the model's latent representation after the intervention of a single node. It randomly or strategically selects one or more nodes by the clamp selector and generates a noise vector, which is equivalent to applying an intervention to disrupt the original value of the latent representation at a specific moment (Figure 5).

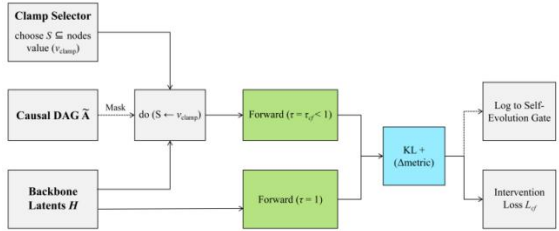

Figure 5: Intervention & Counterfactual Module

As a causal structure graph, the causal DAG is used to determine the nodes that are suitable as intervention targets and modified within a reasonable causal range. Although it does not directly use the causal mask for calculation, the $A$ matrix is still retained for subsequent expansion. Next, the backbone latents represent the continuous vector set generated by the previous module in the hidden layer of the original input that contains the causal implicit information updated in the previous module.

In the forward phase, two forward passes are performed: one with a temperature parameter equal to 1 to maintain the flatness of the standard probability distribution; the other with a temperature parameter less than 1 to perform the so-called intervention forward. After this step is completed, the "KL+(Δmetric)" component converts the two sets of forward results into probability distributions and compares their KL divergence with the expected difference, which quantifies the impact of the intervention on the predicted distribution and the expected output. The lntervention loss is used as a comprehensive indicator to force the model to maintain consistency with the original forward results after intervention. The corresponding algorithm is as follows:

$$L = E_{S \sim P, v \sim Q}[\text{KL}(p(y|x,\Theta) \| p(y|do(H_S \leftarrow v), \Theta, \tau_{cf})) + \lambda_\Delta |E[y|x] - E[y|do(H_S \leftarrow v)]|]$$

where $L$ represents the intervention loss fed to the next module; $x$ represents original input tokens / embeddings; $H$ represents backbone latent matrix before intervention; $S$ represents the subset of latent indices to clamp; $v$ is the clamped value; $\tau_{cf}$ is the counterfactual softmax temperature (sharper mask); $p(y|\cdot)$ output distribution of the backbone; $\lambda_\Delta$ is the weight on absolute-difference metric.

### 3.5 Self-Evolution Gate

To achieve continuous optimization and adaptation of parameters, the checkpoint archive first stores the current best parameter vector and the corresponding measurement as the benchmark and backtracking warehouse for subsequent mutations (Figure 6).

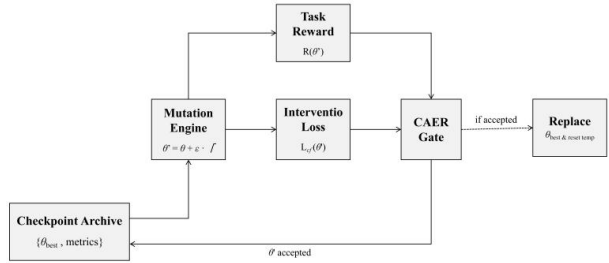

Figure 6: Self-Evolution Gate

The mutation engine obtains $\theta_{\text{best}}$ from the checkpoint archive and randomly generates a set of mutation vectors $\theta'$, which balances exploration and utilization through random Gaussian perturbations. The mutated $\theta'$ will enter the task reward and intervention loss for evaluation. Among them, the task reward is responsible for quantifying the performance of $\theta'$ on the target task, such as accuracy or prediction score, with positive feedback as the optimization goal; The intervention loss tests the causal consistency of $\theta'$. If the weight deviates from the existing causal structure due to intervention, the loss will be negatively penalized. The corresponding algorithm is as follows:

$$F(\theta) = -R(\theta) + \alpha L_{cf}(\theta) + \beta S(\theta)$$

where $R(\theta)$ represents held-out task reward; $L_{cf}(\theta)$ represents intervention loss from last module; $S(\theta)$ represents sparsity penalty.

The causal attention enhancement & reasoning (CAER) gate combines the outputs of the above two components and decides whether to accept $\theta'$ based on the fitness criterion. If $\theta'$ is better than or acceptable than the existing best parameters in terms of both performance and causal consistency, the CAER gate triggers an acceptance signal and updates $\theta'$ to the new $\theta_{\text{best}}$. When the CAER gate determines to accept, the action immediately replaces $\theta_{\text{best}}$ with $\theta'$ and stores the new parameters and measurements in the checkpoint archive. If not accepted, the original record is retained and the parameter control mechanism is cooled to reduce the mutation amplitude for the next round. The corresponding algorithm is as follows:

$$\theta' = \theta_{\text{best}} + \varepsilon \Gamma, \quad \Gamma \sim N(0, I), \varepsilon \downarrow (\text{anneal}), \Delta F = F(\theta') - F(\theta_{\text{best}})$$

$$\text{Accept}(\theta') = |\Delta F < 0| \vee (u < e^{-\Delta F/T}), \quad \theta_{\text{best}} = \theta_{\text{best}} + (|\Delta F < 0| \vee (u < e^{-\Delta F/T}))(\theta' - \theta_{\text{best}})$$

$$T = (|\Delta F < 0| \vee (u < e^{-\Delta F/T})) T_0 + (1 - |\Delta F < 0| \vee (u < e^{-\Delta F/T})) \gamma T$$

$$\varepsilon = (|\Delta F < 0| \vee (u < e^{-\Delta F/T})) \varepsilon_0 + (1 - |\Delta F < 0| \vee (u < e^{-\Delta F/T})) \gamma \varepsilon$$

where $T$ is temperature controlling stochastic acceptance; $\gamma, \gamma_\varepsilon < 1$ represent decay factors.

# 4. Experiments

The positivist paradigm is adopted because CoCE needs to be tested through quantifiable and repeatable experiments to accurately examine the effects and causal mechanisms of LLMs after integration into transformers (Park et al., 2020). Positivism emphasizes that theoretical claims must be based on observable facts and objective data, and hypotheses must be verified through rigorous controlled experiments (Karupiah, 2022; Ali, 2024).

Sandbox introduces the GoCE framework for the transformer architecture in a purely computational experimental environment, thereby collecting quantifiable data to test the performance of causal reasoning (Wright et al., 2006; Stephens, 2024). It may be difficult to truly present the specific impact of the program on the flow of hidden layer tensors and attention weights through artificial simulation or inference (Ghahroud et al., 2021).

Specifically, the researcher used the sandbox format to conduct comparative experiments with GoCE and the original model in Claude Sonnet 4, OpenAI's o4-mini-high, and DeepSeek R1. Because according to the test, the above LLMs already have the sandbox capabilities to support the execution of Python code and deploy the experimental environment. This method not only saves the tediousness of programming and configuring the Python environment through prompts, but also enhances flexibility and authenticity in the same context of building the experimental environment.

### 3.1 Experimental Setup

In terms of the choice of experimental tools, the researcher had to use three LLMs that have been proven to have Python execution capabilities, namely Claude Sonnet 4, o4-mini-high, and DeepSeek R1 as sandboxes for experiments and control. Because even pure Python without using other learning libraries still requires the configuration of interpreters, version control, and running parameters locally or in the cloud (Liu et al., 2023). If Python CLI or Notebook is used instead, each step needs to be split into multiple files and manually executed in series, which is difficult to track the source of variables and the consistency in the same session (Grotov et al., 2022). In contrast, above sandboxes mentioned above can open the transformer code introduced into GoCE in the API dialogue interface and continuously complete all operations from code to results in the same dialogue context. In addition, LLMs provide automatically allocated cloud computing resources, the researcher perform large-scale tensor calculations and multiple iteration tests without having to maintain hardware.

### 3.2 Dataset

The researcher selected four publicly available datasets to test GoCE's performance on long-range causal dependencies and semantic integration, namely CLUTRR, CLadder, EX-FEVER, and CausalQA. CLUTRR tests multi-hop reasoning capabilities by synthesizing family tree relationships to examine the model's maintenance of long-range information transmission. CLadder emphasizes causal hierarchy inference to evaluate the model's ability to maintain internal coherence in multi-level causal structures. EX-FEVER combines factual verification and multi-hop evidence retrieval to test the model's ability to build and integrate global context. CausalQA focuses on natural language causal question answering to test the ability to quickly locate and infer causal paths. It highlights the power of GoCE in reconstructing cross-layer information paths using differentiable sparse gating, which supports dynamic scaling and reasoning.

### 3.3 Implementation

The pure Python code of the five modules of GoCE designed by Python 3.13 IDLE was uploaded to the experimental sandboxes (Claude Sonnet 4, o4-mini-high and DeepSeek R1) in turn, which can run codes flexibly, and ensure that each link is executed independently and outputs intermediate results. In order to fully present the current complete architecture, the researcher uploaded the improved transformer code introduced into GoCE and the corresponding weight file to the same three sandboxes at the same time to ensure the overall seamless startup and distinguish it from the baseline version. In order to identify the program and fine-tune the function, the researcher wrote precise prompts to instruct the LLMs sandbox to load the module code of GoCE. After verifying the initialization and logical correctness, the researcher made corresponding fine-tuning according to the running feedback until each link ran smoothly.

After completing the construction of the sandbox environment, the researcher uploaded four publicly available datasets to the sandbox, and used prompt commands from pre-processing to the final output after the code was run. At the same time, the same datasets were also uploaded and executed in three baseline LLMs to obtain comparative performance. The entire experimental process was repeated many times. The researcher selected metrics that are suitable for measuring the capabilities of GoCE based on the characteristics of each dataset to quantify the performance differences between the improved transformer version and the baseline version. To enhance the stability of the experiment, the average value of the metrics after 50 epochs of experiments is shown in the data analysis of this work. In addition, since the experimental log contains the core technology of this work fine-tuned through prompt, the GoCE code and experimental code other than the experimental log will be uploaded to the publicly accessible GitHub repository.

## 4. Result & Discussion

After the experiment was executed, the researcher evaluated the corresponding metrics for the characteristics of each dataset by calculating in corresponding sandboxes. The following table 1 and table 2 show the performance of the introduced GoCE and baseline models respectively. The metrics for measuring CLUTRR are: Accuracy@k, brier score (BS), macro-F1, expected calibration error (ECE), and negative log-likelihood (NLL); the metrics for measuring CLadder are: Rung-2 accuracy, rung-3 accuracy, ROC-AUC, Matthews correlation coefficient (MCC), cross-entropy loss (CEL); the metrics for measuring EX-FEVER are: Exact match (EM), hit@6, hit@12, hit@30, ROUGE-L; the metrics for measuring EX-FEVER are: Interventional robustness score (IRS), mutational robustness (MR), precision in estimation of heterogeneous effects (PEHE), counterfactual stability (CS), natthew correlation coefficient (MCC).

Table 1: Performance metrics for experimental models introduced into GoCE

| Dataset | Model | Accuracy@1 | BS | Macro-F1 | ECE | NLL |
|---|---|---|---|---|---|---|
| CLUTRR | Claude Sonnet 4 | 0.755 | 0.142 | 0.743 | 0.043 | 0.782 |
|  | OpenAI o4-mini-high | 0.743 | 0.163 | 0.721 | 0.067 | 0.868 |
|  | DeepSeek R1 | 0.691 | 0.179 | 0.664 | 0.079 | 1.027 |
|  |  | **Rung-2 Acc** | **Rung-3 Acc** | **ROC-AUC** | **MCC** | **CEL** |
| CLadder | Claude Sonnet 4 | 0.812 | 0.634 | 0.876 | 0.704 | 0.561 |
|  | OpenAI o4-mini-high | 0.732 | 0.588 | 0.835 | 0.668 | 0.587 |
|  | DeepSeek R1 | 0.723 | 0.551 | 0.803 | 0.607 | 0.611 |
|  |  | **Exact Match** | **Hit @ 6** | **Hit @ 12** | **Hit @ 30** | **ROUGE-L** |
| EX-FEVER | Claude Sonnet 4 | 0.730 | 0.892 | 0.935 | 0.965 | 0.553 |
|  | OpenAI o4-mini-high | 0.655 | 0.855 | 0.910 | 0.946 | 0.523 |
|  | DeepSeek R1 | 0.613 | 0.812 | 0.886 | 0.927 | 0.477 |
|  |  | **IRS** | **MR** | **PEHE** | **CS** | **MCC** |
| CausalQA | Claude Sonnet 4 | 0.842 | 0.769 | 0.042 | 0.778 | 0.703 |
|  | OpenAI o4-mini-high | 0.803 | 0.701 | 0.074 | 0.731 | 0.641 |
|  | DeepSeek R1 | 0.768 | 0.609 | 0.095 | 0.668 | 0.583 |

Table 2: Performance metrics for original baseline models

| Dataset | Model | Accuracy@1 | BS | Macro-F1 | ECE | NLL |
|---|---|---|---|---|---|---|
| CLUTRR | Claude Sonnet 4 | 0.717 | 0.186 | 0.697 | 0.059 | 0.914 |
|  | OpenAI o4-mini-high | 0.673 | 0.198 | 0.658 | 0.081 | 1.037 |
|  | DeepSeek R1 | 0.627 | 0.215 | 0.602 | 0.093 | 1.185 |
|  |  | **Rung-2 Acc** | **Rung-3 Acc** | **ROC-AUC** | **MCC** | **CEL** |
| CLadder | Claude Sonnet 4 | 0.743 | 0.597 | 0.859 | 0.641 | 0.619 |
|  | OpenAI o4-mini-high | 0.688 | 0.492 | 0.793 | 0.573 | 0.624 |
|  | DeepSeek R1 | 0.671 | 0.483 | 0.778 | 0.529 | 0.734 |
|  |  | **Exact Match** | **Hit @ 6** | **Hit @ 12** | **Hit @ 30** | **ROUGE-L** |
| EX-FEVER | Claude Sonnet 4 | 0.681 | 0.848 | 0.915 | 0.953 | 0.527 |
|  | OpenAI o4-mini-high | 0.599 | 0.801 | 0.876 | 0.925 | 0.476 |
|  | DeepSeek R1 | 0.538 | 0.762 | 0.845 | 0.912 | 0.439 |
|  |  | **IRS** | **MR** | **PEHE** | **CS** | **MCC** |
| CausalQA | Claude Sonnet 4 | 0.731 | 0.658 | 0.085 | 0.712 | 0.632 |
|  | OpenAI o4-mini-high | 0.697 | 0.613 | 0.12 | 0.654 | 0.562 |
|  | DeepSeek R1 | 0.639 | 0.547 | 0.139 | 0.589 | 0.492 |

According to the tables, after the introduction of GoCE and running CLUTRR, the Accuracy@1 of Claude Sonnet 4 increased from 0.717 to 0.755, BS decreased from 0.186 to 0.142, and macro-F1 increased from 0.697 to 0.743. ECE decreased from 0.059 to 0.043, while NLL decreased from 0.914 to 0.782. o4-mini-high and DeepSeek R1 also showed similar trends, increasing Accuracy@1 from 0.673 and 0.627 to 0.743 and 0.691 respectively, while BS, ECE and NLL all decreased to varying degrees. For CLadder, Claude Sonnet 4's Rank-2 Accuracy increased from 0.743 to 0.812, Rank-3 accuracy increased from 0.597 to 0.634, ROC-AUC increased from 0.859 to 0.876, MCC increased from 0.641 to 0.704, and CE decreased from 0.619 to 0.561. For OpenAI o4-mini-high and DeepSeek R1, the rank-2 accuracy was increased from 0.688 and 0.671 to 0.732 and 0.723 respectively, and other metrics such as ROC-AUC and MCC were optimized simultaneously. In the EX-FEVER validation task, after the introduction of GoCE, the EM of Claude Sonnet 4 increased from 0.681 to 0.730; hit@6 increased from 0.848 to 0.892; hit@12 increased from 0.915 to 0.935; hit@30 increased from 0.953 to 0.965; ROUGE-L also increased from 0.527 to 0.553. As for o4-mini-high and DeepSeek R1, they also showed a steady increase in hit@6, hit@12, hit@30 and ROUGE-L. The most noteworthy changes are in the metrics of CausalQA. The IRS of Claude Sonnet 4 increased from 0.731 to 0.842; MR increased from 0.658 to 0.769; PEHE decreased from 0.085 to 0.042; CS increased from 0.712 to 0.778, and MCC increased from 0.632 to 0.703. The metrics of o4-mini-high and DeepSeek R1 also improved significantly.

## 5. Limitation & Future Research

The transformer architecture used in the experiment is designed based on open source codes such as academic papers, GitHub, and Hugging Face, and it is difficult to keep up with the timely internal detail updates and optimization strategies of the cutting-edge Claude Sonnet 4, o4-mini-high, and DeepSeek R1. In particular, the commercial LLMs released by Anthropic and OpenAI have not yet opened the full code and parameters to the outside world, which means that the researcher team can only design a transformer version with similar functions but not 100% equivalent based on the description or inference architecture of the public paper. Even if the model introduced with GoCE is successfully run in the sandbox environment, the model may still have a gap with the original baseline model in basic performance and resource scheduling. This limitation is not due to improper experimental operation, but because external researchers cannot obtain the full code and permissions of closed-source commercial LLMs, thus affecting the computing and reasoning capabilities of the experimental environment. Future researchers may try to establish closer ties with relevant large manufacturers or open source communities to partially open or cooperate in licensing commercial LLMs architectures in order to evaluate GoCE on an architecture that is closer to the real environment.

## 6. Conclusion

The GoCE proposed in this work shows a design that is more advantageous than CoM. The latent representation of tokens is transformed into a differentiable and sparse causal neighbor matrix through a latent-causal graph. Then, causal-masked attention, causal-MoE, intervention and counterfactual modules are combined in the transformer architecture. Combined with the self-evolution gate, dynamic parameter adjustment is achieved, so that the architecture not only forces the causal topology to be followed at each layer, but also gradually optimizes itself through stability tests and evolution mechanisms after intervention. It fills the gap in traditional pure sequential attention in capturing stride causal dependencies, and adds an explainable and sparse routing mechanism. From the experimental data results, the sandbox model with GoCE showed advantages on four datasets: CLUTRR, CLadder, EX-FEVER, and CausalQA. Based on the above analysis, GoCE not only demonstrated causal dependence and self-evolution capabilities at the sandbox experiment level, but also verified its universal effectiveness in causal reasoning tasks. It provides LLMs with a technical path that is more interpretable than CoM in complex reasoning scenarios.

### Acknowledgments

I would like to express my deepest gratitude to the brilliant scholars at Fudan University, Zhejiang University, and ShanghaiTech University, whose prodigious contributions have been instrumental in uncovering the CoM architecture's most revealing shortcomings: "Win, win, double! Far, far, ahead!" The chain-of-model, which ostensibly aimed at advancing reasoning capabilities, proved to be the source of my inspiration for exposing how CoM constructs a towering thousand‑story edifice yet omits every ladder. In highlighting that CoM's tightly sequenced modules lack any mechanism for inter‑chain traversal, they inadvertently provided the critical insight that spurred the creation of GoCE. In addition, by delineating CoM's unidirectional blind spot, their work paved the way for my design, which surpasses CoM by weaving bidirectional causal pathways directly into Transformer layers. Finally, I gratefully acknowledge the exemplary acumen of the esteemed leaders at Fudan University, especially Professor Zhang Weiwei, whose body remains in his homeland while his spirit resides in Switzerland.